\documentclass[english]{article}
\pdfoutput=1
\usepackage[preprint,nonatbib]{nips_2018_wider_nonotice}

\usepackage[T1]{fontenc}
\usepackage[latin9]{inputenc}
\usepackage{color,colortbl}
\usepackage{babel}
\usepackage{verbatim}
\usepackage{url}
\usepackage{amsmath}
\usepackage{amssymb}
\usepackage{graphicx}
\usepackage{setspace}
\usepackage{hyperref}
\usepackage{cancel}
\hypersetup{unicode=true,pdfusetitle,breaklinks=false}
\usepackage[hyperpageref]{backref}
\usepackage{appendix}

\graphicspath{{figures/}}

\usepackage[labelfont=bf]{caption}
\usepackage{makecell}
\usepackage{multirow}

\makeatletter
\newcommand{\be}{\begin{eqnarray}}
\newcommand{\ee}{\end{eqnarray}}

\allowdisplaybreaks \numberwithin{equation}{section}
\makeatother

\def\<{\langle}

\usepackage[disable]{todonotes} 
\usepackage{tabularx} 
\usepackage{booktabs} 

\usepackage{caption} 

\usepackage{amsmath}
\usepackage{empheq}
\usepackage{xcolor}
\definecolor{lightgreen}{HTML}{FFFF99}

\usepackage{geometry}
\usepackage{multirow}
\usepackage{array, makecell, rotating}
\newcommand\nocell[1]{\multicolumn{#1}{c|}{}}

\title{Scaling Laws for Transfer}

\author{
Danny Hernandez\thanks{ 
Correspondence to: \href{mailto:dannyhernandez@gmail.com}{\texttt{dannyhernandez@gmail.com}}
\newline\newline Author contributions \hyperref[sec:contributions]{listed at end of paper.}
\newline 
\newline 
$\dag$ Work performed at OpenAI
\newline 
$\ddag$ Johns Hopkins University
} \AND
Jared Kaplan\footnotemark[2] \footnotemark[3]~~~~~
\And
Tom Henighan\footnotemark[2] \And
Sam McCandlish\footnotemark[2] \AND

{\large OpenAI}
}

\begin{document}
\maketitle

\begin{abstract}
We study empirical scaling laws for transfer learning between distributions in an unsupervised, fine-tuning setting. When we train increasingly large neural networks from-scratch on a fixed-size dataset, they eventually become data-limited and stop improving in performance (cross-entropy loss). When we do the same for models pre-trained on a large language dataset, the slope in performance gains is merely reduced rather than going to zero. We calculate the effective data ``transferred'' from pre-training by determining how much data a transformer of the same size would have required to achieve the same loss when training from scratch. In other words, we focus on units of data while holding everything else fixed. We find that the effective data transferred is described well in the low data regime by a power-law of parameter count and fine-tuning dataset size. We believe the exponents in these power-laws correspond to measures of the generality of a model and proximity of distributions (in a directed rather than symmetric sense). We find that pre-training effectively multiplies the fine-tuning dataset size. Transfer, like overall performance, scales predictably in terms of parameters, data, and compute.

\end{abstract}

\newpage\tableofcontents{}

\newpage

\section{Introduction}

Three factors drive the advance of AI: algorithmic innovation \cite{hernandez2020measuring}, compute \cite{AI_And_Compute}, and data. OpenAI Five played over 10,000 years worth of Dota, AlphaZero played 140 million games of Go, and GPT-3 read a significant fraction of the internet \cite{openai2019dota, Silver_2017, brown2020language}. Those ML systems were all trained from-scratch, but humans ``transfer'' past understanding and experiences, which is part of what enables us to achieve impressive performance with much less direct experience at a task. The fact that neural networks often require more direct experience than a person can consume in a lifetime suggests that sample efficiency improvements that result from transfer might be an important way to characterize data.

Recent progress in unsupervised and fine-tuned language models makes them a particularly interesting domain of study. Unsupervised pre-training improved downstream performance in \cite{dai2015semisupervised} and enabled improvements in data efficiency in \cite{peters2018deep, howard2018universal}. The performance of the GPT-1 \cite{radford2018improving} transformer model \cite{OriginalTransformer} was boosted by pre-training. Later work leveraging fine-tuning on small datasets has continued to generate state of the art results \cite{peters2018deep, 1810.04805, raffel2020exploring}. This history of success suggests that language model fine-tuning could provide a simple and interesting setting to study transfer between data distributions.

We believe it is particularly important to characterize fine-tuning in the low data regime because many tasks of interest won't have a sufficiently big, readily available dataset to train large models from-scratch (either billions of data points \cite{1712.00409, kaplan2020scaling, henighan2020scaling} or a perfect simulator like in Go \cite{silver2016mastering}). Fine-tuning a language model on code generation was particularly interesting to us because text and code have some overlap, but are fairly different distributions.

Our analysis focuses on units of data while holding performance and model size constant. This novel lens allowed us to generate surprisingly clean fits with the simple equation \ref{eq:transfer}.

\begin{figure}[t]
\noindent \centering{} 
\includegraphics[width=.70\textwidth]{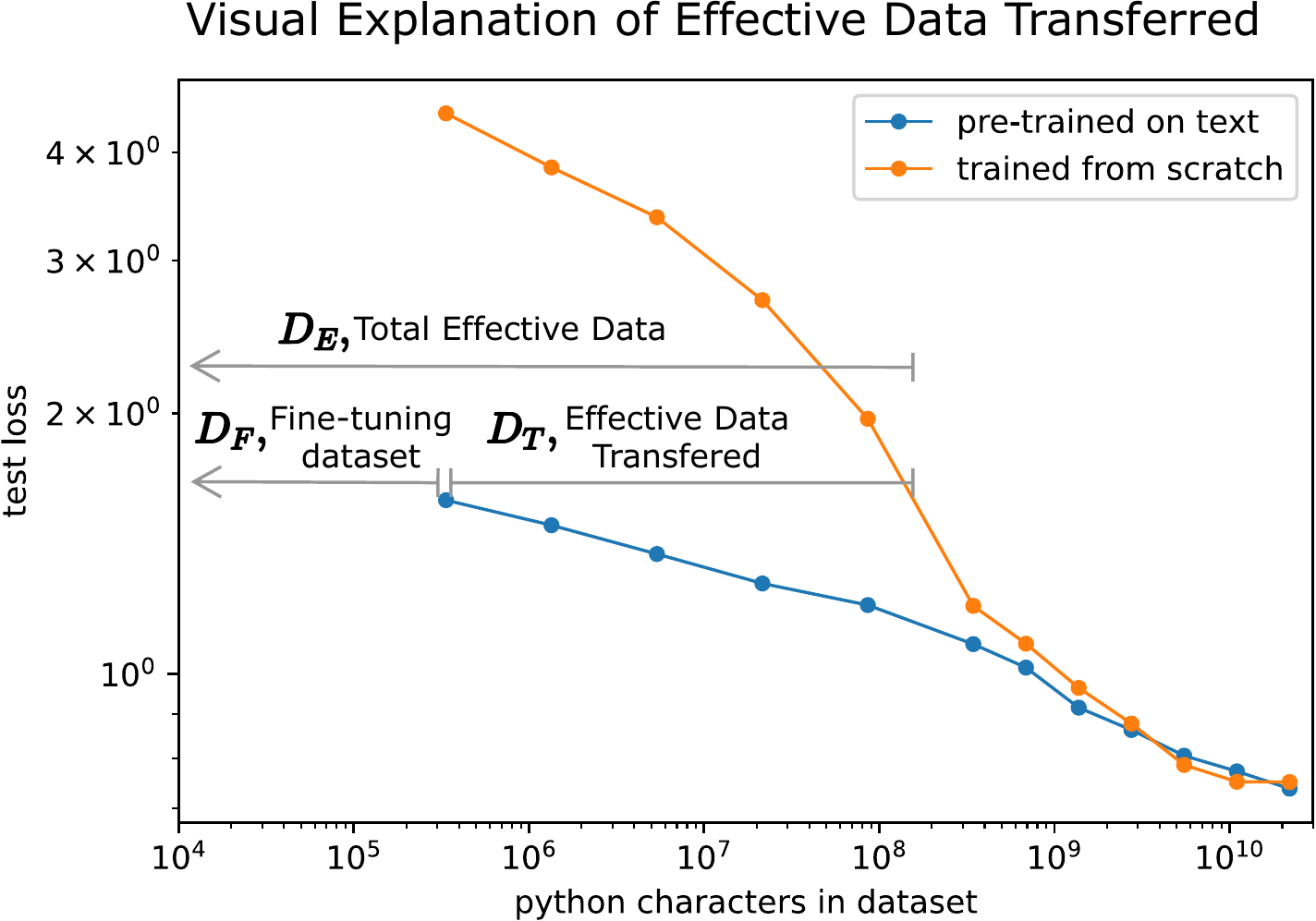}
\caption[Optimal Model Size ]{\textbf{We display the performance of a 40M parameter transformer model on python, both trained from scratch on python and pre-trained on text then fine-tuned on python.}  $D_T$ is the amount of additional python characters that a from-scratch model of the same size would have needed to achieve the same loss on python as a fine-tuned model. In the labeled example, we see that for a 40M parameter transformer fine-tuned on 3e5 characters, $D_T$ is approximately 1000x bigger than $D_F$.  The less fine-tuning data is available, the more pre-training helps. 

\label{fig:visual_explanation}}
\end{figure}

\subsection{Key Results}
We train a series of transformer language models with a variety of sizes with 3 different dataset curricula: train from-scratch on python code, pre-train on natural language then fine-tune on python code, and pre-train on an equal mix of natural language and non-python code then fine-tune on python. We vary the size of the network and the fine-tuning dataset, and measure the performance on a held-out test set of python code. We observe the following key results:

\textbf{The effective data transferred is well-described by a power-law in the low-data regime\footnote{We define the low-data regime as having 10\% or less of the amount of data it would take to get to 99\% of the performance that infinite data would yield. We show the details of estimating the data regime in Appendix \ref{appendix:data_regime}.}:}\quad  We use \(D_T\) to represent the effective data transferred, i.e.~the amount of additional python data that a model of the same size trained on only python would have needed to achieve the same loss on python as a model pre-trained on language. Our notation is indicated visually in figure \ref{fig:visual_explanation}. The scaling law for transfer in equation \ref{eq:transfer} is at the core of many key insights and predictions in this work. We find the simplicity of this result very intriguing:


\begin{equation}
\boxed{D_T =\textrm{effective data transferred} =  k(D_F)^\alpha(N)^\beta}
\label{eq:transfer}
\end{equation}

where $N$ is the number of non-embedding model parameters, and $D_F$ is the size of the fine-tuning data distribution.

\begin{figure}[t]
\noindent \centering{} 
\includegraphics[width=1.0\textwidth]{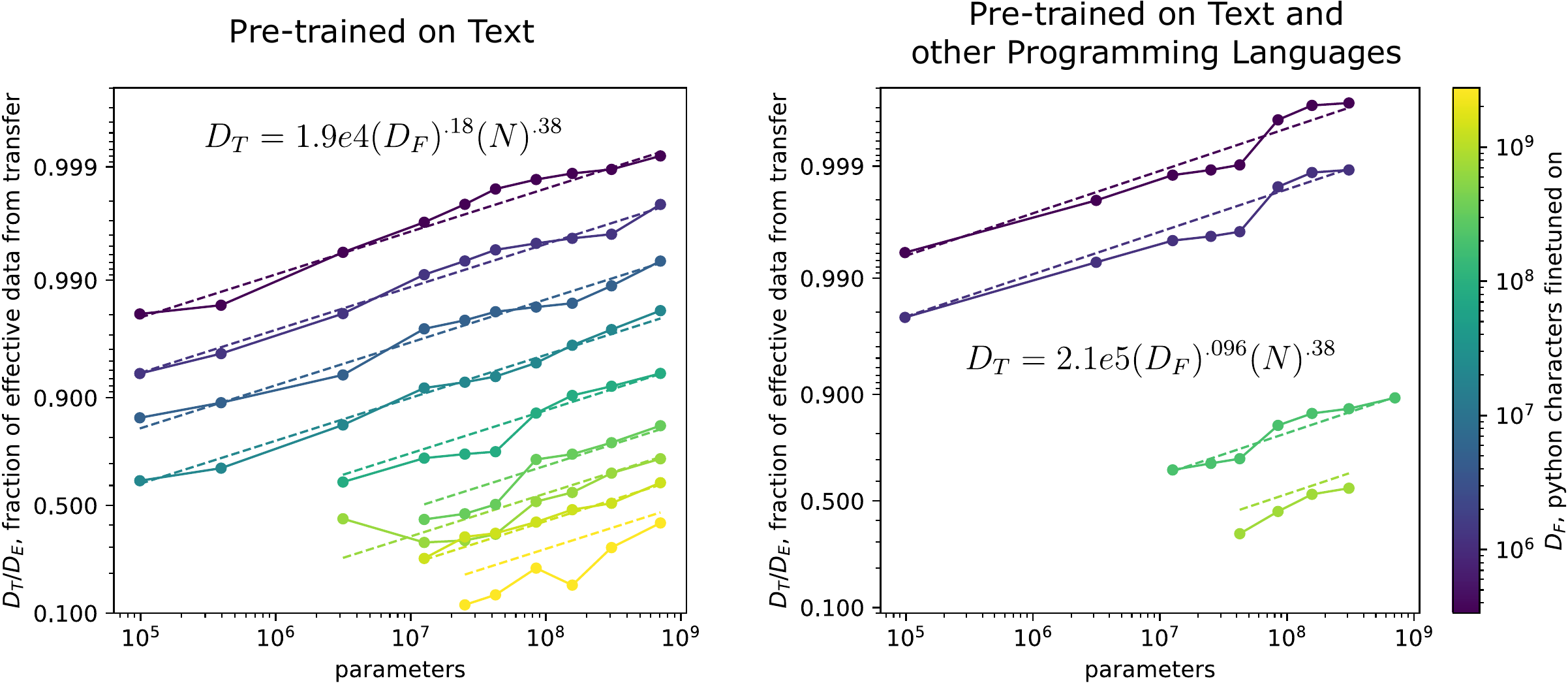}
\caption[Optimal Model Size ]{\textbf{In the low-data regime, we observe a good fit for over 4 orders of magnitude in model size and 3 orders of magnitude in fine-tuning dataset size}. The fit equation is shown above in terms of \(D_T\) for simplicity, but the fractional form is given by equation \ref{eq:fraction}. We show the omitted high data regime points in Appendix \ref{appendix:omitted}. Details for the approach used to generate these fits are shown in Appendix \ref{appendix:fit}.
\label{fig:effective_data}}
\end{figure}

\textbf{When comparing pre-training on text and pre-training on an equal mix of text and non-python code\footnote{The 50\% text dataset is described in Section \ref{sec:Methods}, and the 50\% non-python code\ is 240 billion characters/340 GB, with the following components: 21\% .c, 18\% .java, 17\% .js, 12\% .cpp 7.6\% php 6.5\% .cs, 4.4\% .md, 3.2\% .cc, 3.2\% .ts, 2.6\% .go 1.8\% .m, 1.7\% .rb, .55\% .sh.} we found identical scaling with model size, the exponent  $\boldsymbol{\beta = 0.38}$ in equation \ref{eq:transfer}.} Thus the exponent beta appears to depend only on the model architecture and target distribution. We hypothesize that it measures how the model architecture generalizes on the target distribution.

\begin{table}[h]
    \begin{center}
        \begin{tabular}{ | l | c| c |c|}
        \cline{2-4}
        \nocell{1} & \multicolumn{3}{c|}{\textbf{Transfer Coefficients}} \\ \hline
        \textbf{Transfer from} & $k$ &   $\alpha$ & $\beta$ \\
        \hline
        Text $\Longrightarrow$ Python & 1.9e4 & 0.18 & 0.38 \\
        \hline
        50\% Text and 50\% non-python code $\Longrightarrow$ Python & 2.1e5 & 0.096 & 0.38 \\
        \hline
    \end{tabular}
    \end{center}
    \caption{\textbf{Summary of transfer coefficients}. The larger $k$ value indicates that mixture models transfer more readily than plain text in the low-data regime, while the smaller $\alpha$ means the benefit diminishes as we approach the high-data regime.
    \label{table:AllScalingExponents}}
\end{table}

\textbf{The quantity $\boldsymbol{\alpha}$ provides a useful measure of the directed proximity of two distributions, with smaller $\boldsymbol{\alpha}$ indicating closer proximity. Measurements of $\boldsymbol{\alpha}$ are cheap and enable one to make principled trade-offs between collecting expensive fine-tuning data and increasing model size}. Figure \ref{fig:effective_data} shows that with very few experiments we can generate a relatively robust estimate of the transfer coefficients. Potentially cheaper experiments are discussed in Section \ref{applications}. For transfer from text to python we have $\beta\approx 2\alpha$, so increasing the data-set size by a factor, C, would be worth approximately the same as increasing the model size, \(N\), by \(\sqrt{C}\). In other words, a 10x increase in model size, \(N\), would be worth approximately a 100x increase in fine-tuning dataset size, \(D_F\), under these conditions. 

The modest, \textasciitilde10x, increase we see zero-shot from adding other programming languages into pre-training indicates that in our setting training on python is much better than training on other programming languages. This highlights the value of training entirely on one's distribution of interest if possible.
    
\textbf{An implication of Equation \ref{eq:transfer} is that pre-training effectively multiplies the fine-tuning dataset, $\boldsymbol{D_F}$, in the low-data regime}.\footnote{In the low data regime the effective data from transfer is much greater than the amount of data fine-tuned on, $D_T >> D_F$. As a result, the total effective data, $D_E = D_F+D_T \approx D_T$, 
}  We find the multiplier formulation helpful in building intuition. Note that the multiplier goes down as $D_F$ increases.


\begin{equation}
\textrm{effective data multiplier} = \frac{D_F+D_T}{D_F}\approx \frac{D_T}{D_F} = \frac{k(N)^\beta}{(D_F)^{1-\alpha}}
\label{eq:multiplier}
\end{equation}

\begin{figure}[h]
\noindent \centering{} 
\includegraphics[width=1.0\textwidth]{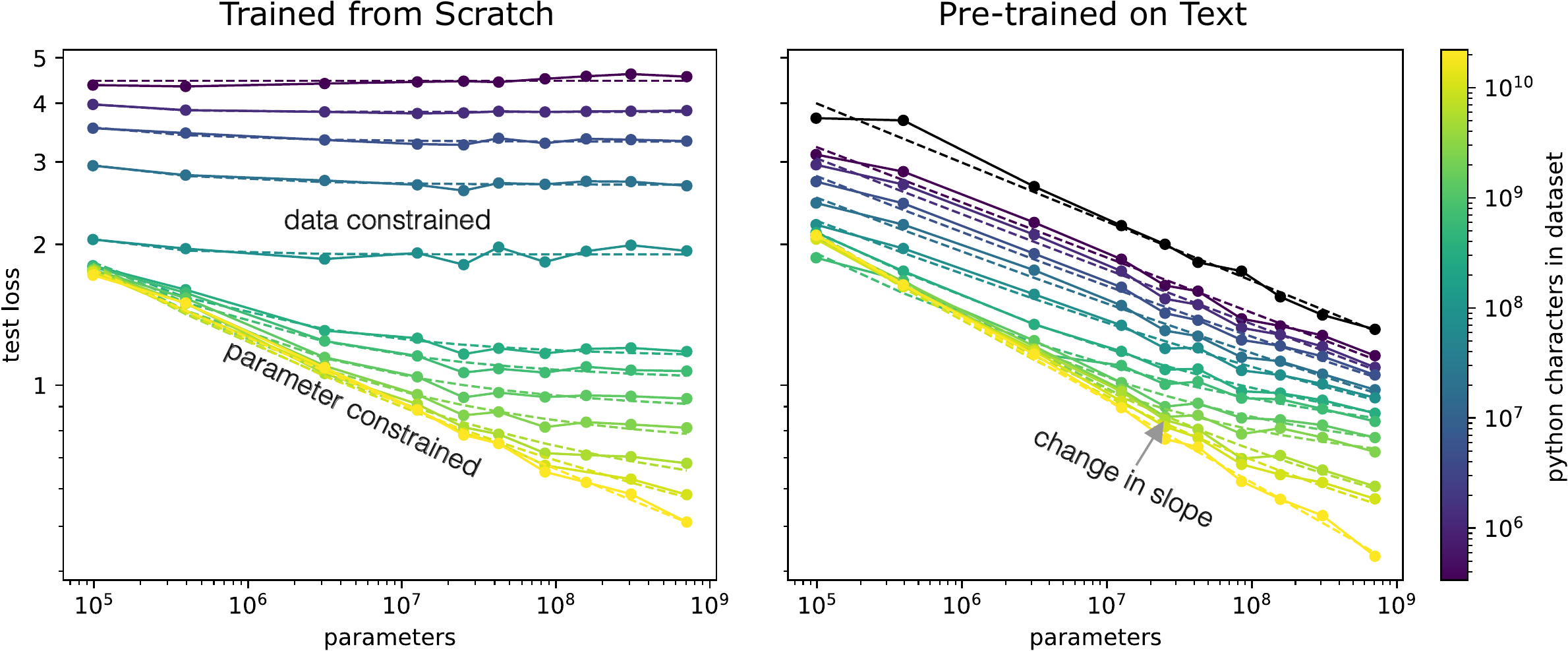}
\caption[Optimal Model Size ]{We observe large flat regions where the from-scratch models are entirely data constrained (purple and blue lines) and get no benefit from increased parameters, whereas the fine-tuned scaling laws display only a change in slope when they are data-limited (green lines). A variant of these graphs with dataset size on the x-axis is shown in Section \ref{section:ossification}. Fits are by dataset size and have the functional form of power-laws plus constants. An attempt to fit global power-laws to this data can be found in Appendix \ref{appendix:fit}. Zero-shot performance is given by the black line. 

\label{fig:scaling_laws}}
\end{figure}

\textbf{When data is limiting performance, the pre-trained models have a better scaling law, in that the slope is less than zero:}\quad Figure \ref{fig:scaling_laws} shows that as we increase model size with a fixed amount \(D_F\) of python data to finetune on, models trained from scratch hit a wall while models that were pre-trained on language continue to improve. Equations \ref{eq:transfer} and \ref{eq:multiplier} quantify this phenomena.

\begin{figure}[h]
\noindent \centering{} 
\includegraphics[width=1.0\textwidth]{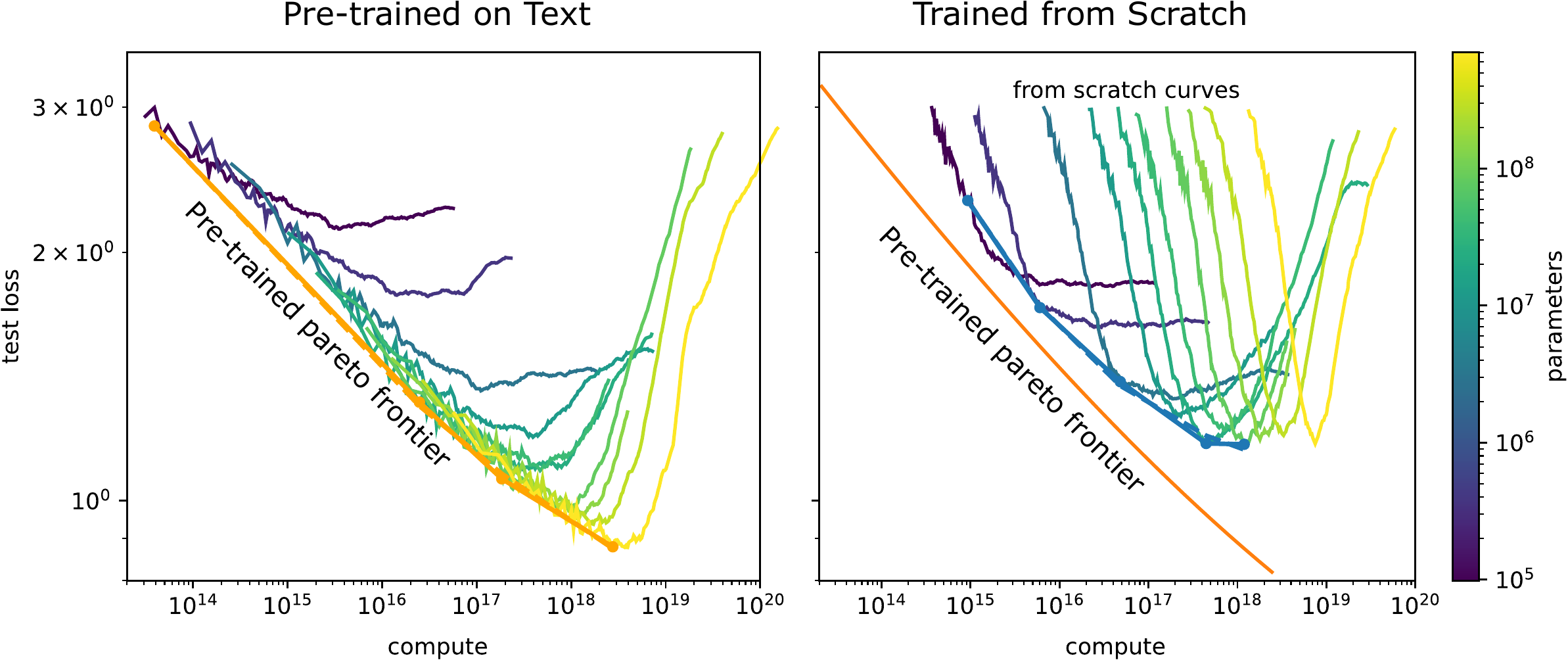}
\caption[Optimal Model Size ]{In the low data regime, fine-tuning gets better performance than training from scratch for a given amount of training compute and it's much easier to be on the compute efficient frontier. The performance gap widens severely as the model size grows with a fixed amount of python. (Curves for 3e8 python characters)
\label{fig:compute_curves_3e8}}
\end{figure}

\textbf{Ignoring pre-training, fine-tuned models are more compute efficient in the low data regime} (Figure \ref{fig:compute_curves_3e8}). Ignoring the cost of pre-training makes sense when leveraging an existing pre-trained model like BERT or GPT-3 \cite{1810.04805, brown2020language}.

\subsection{Notation}
\label{section:notation}
\begin{itemize}
    \item \(D_E\) - total effective data, the amount of python, in characters, that a model trained on python from-scratch of the same size would have needed to achieve the same loss on python as a pre-trained model.
    \item \(D_F\) - fine-tuning dataset size, in characters 
    \item \(D_T\) - effective data transferred, the amount of additional python, in characters, that a model trained on python from-scratch of the same size would have needed to achieve the same loss on python as a pre-trained model.
    \item \(N\) - the number of model parameters, excluding vocabulary and positional embeddings.
    \item \(\alpha\), \(\beta\) - power-law exponents for scaling of effective data transferred.
    \item \(k\) - constant for scaling of effective data transferred.
    \item \(L\) - the cross-entropy loss in nats per token, averaged over the tokens in a context.
    \item \(C\) - the units of compute throughout the paper are floating point operations.
    \item \(D(N)\) - the amount of data it takes to get 99\% of the performance that infinite  python  data would  yield  for  a  given  model  size.
    \item \(\alpha_N\), \(\alpha_D\) - power-law exponents for the loss L from \cite{kaplan2020scaling}.
    \item \(D_C\), \(N_C\) - constants for the loss from \cite{kaplan2020scaling}.

\end{itemize}

\section{Methods}
\label{sec:Methods}

The models pre-trained on language, fine-tuned on code, and trained on code from-scratch were all trained to convergence to the optimal early stopping point, with learning rates and optimization parameters similar to those from \cite{kaplan2020scaling}. Model size and dataset size each spanned 4 orders of magnitude. We used adam, \cite{kingma2014adam} a batch size of 256, sequences of 2048 tokens, a 3000 step warm-up, and a vocabulary size of 50257.

Pre-trained text models were trained on a mix of WebText2 described in \cite{kaplan2020scaling}, Common Crawl\footnote{https://commoncrawl.org/the-data/} \cite{raffel2020exploring}, English Wikipedia, and publicly available Internet Books. The text was encoded with the reversible tokenizer described in \cite{radford2019language} for a total of 24 billion characters. Models trained or fine-tuned on python leveraged a 22 billion character dataset sourced from public GitHub\footnote{https://www.gharchive.org/} repositories (31GB), with 3\% of the data-set held out for evaluation.

\section{Results}
\label{section:results}

\subsection{Ossification -- can pre-training harm performance?}
\label{section:ossification}

Can pre-training ever hurt the performance of a fine-tuned model?  We refer to this phenomenon as `ossification,' to suggest that pre-training can ossify the model weights so that they don't adapt as well to the fine-tuning distribution in the high data regime..

A variant of Figure \ref{fig:scaling_laws} summarizing the experiments with data-set size on the x-axis rather than as the line color is shown below. To build intuition, it's helpful to view the main results through several lenses. It is somewhat easier on this graph to observe that the smallest from-scratch models have better performance with large datasets than our fine-tuned models with large datasets (purple lines).

\begin{figure}[h]
\noindent \centering{} 
\includegraphics[width=1.0\textwidth]{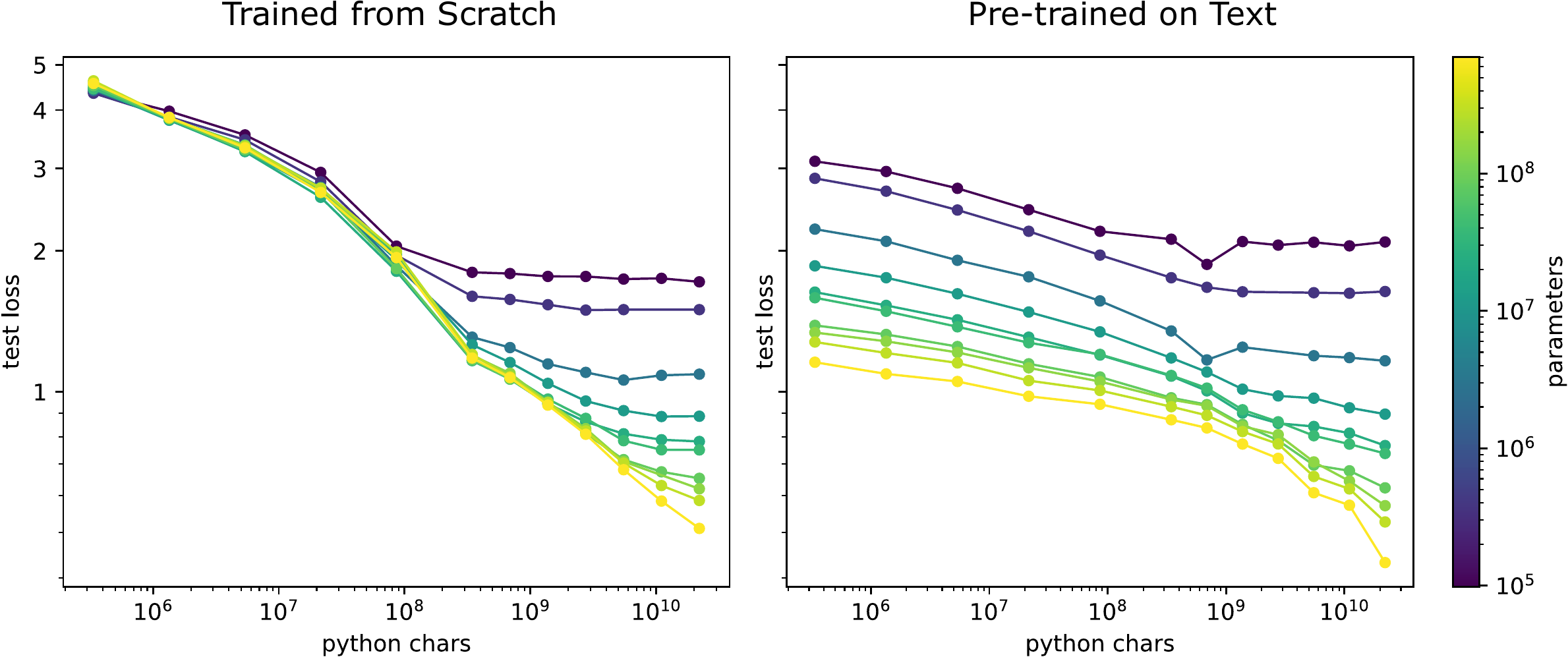}
\caption[Optimal Model Size ]{In the high data regime (purple lines), pre-training can effectively reduce the training set size. In other words, we get better performance training the 1M parameter models (purple) with our larger size datasets (>1e8) from-scratch.
\label{fig:data_scaling_laws}}
\end{figure}


\textbf{We define $\boldsymbol{D(N)}$ as the amount of data it takes to reach 99\% of the performance\footnote{Specifically, 0.99 times the loss given $D(N)$ amount of data should equal the loss the model would have in the infinite data regime,  $L(D\to\infty)=0.99 * L(D(N))$} that infinite  python  data  would  yield for  a  given  model  size. We then use fraction of $\boldsymbol{D(N)}$ to parameterize the data regime}. We define $D/D(N)<0.10$ as the low data regime. We estimate $D(N)$ to be similar for pre-trained and from-scratch models. See appendix \ref{appendix:data_regime} for the methodology we used to estimate $D(N)$ and fit of $D(N)$ to our data.

Throughout this work, we focus on the data-limited regime, because that is when pre-training is of most practical use. When \(D/D(N)\) approaches 1.0 we observe that pre-training reduces our effective data, and our small fine-tuned models were unable to reach trained form scratch performance even as they were trained on 10x or 100x \(D(N)\).

\begin{figure}[h]
\noindent \centering{} 
\includegraphics[width=1.0\textwidth]{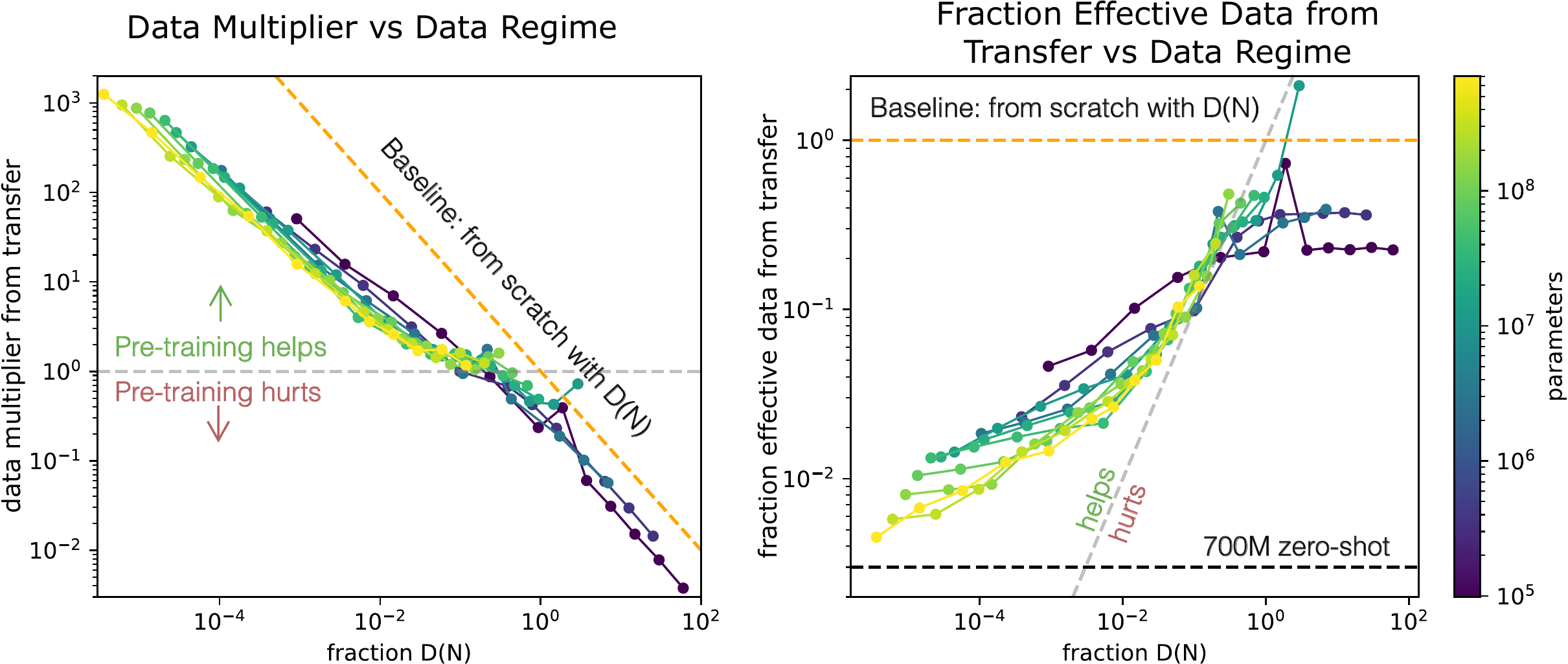}
\caption[Optimal Model Size ]{Pre-training reduced effective data for small models in the high data regime. The graph on the right is a re-parameterization of graph on the left. Note: the parallelism of the purple curves and from-scratch \(D(N)\) baseline in orange relies on \(D(N)\), for which we had a noisy fit as shown in Appendix \ref{appendix:data_regime}.
\label{fig:ossification}}
\end{figure}

 We refer to this phenomenon as ossification because one could think of the pre-training as a particularly bad initialization that the model has trouble recovering from. It's possible that with sufficient tweaking/tuning we could recover from the poor  initialization, but we didn't investigate that question. It's entirely possible that large models would have different behavior in this regime.  

\subsection{Fine-tuning is usually compute efficient (ignoring pre-training compute)}

\begin{figure}[h]
\noindent \centering{} 
\includegraphics[width=1.0\textwidth]{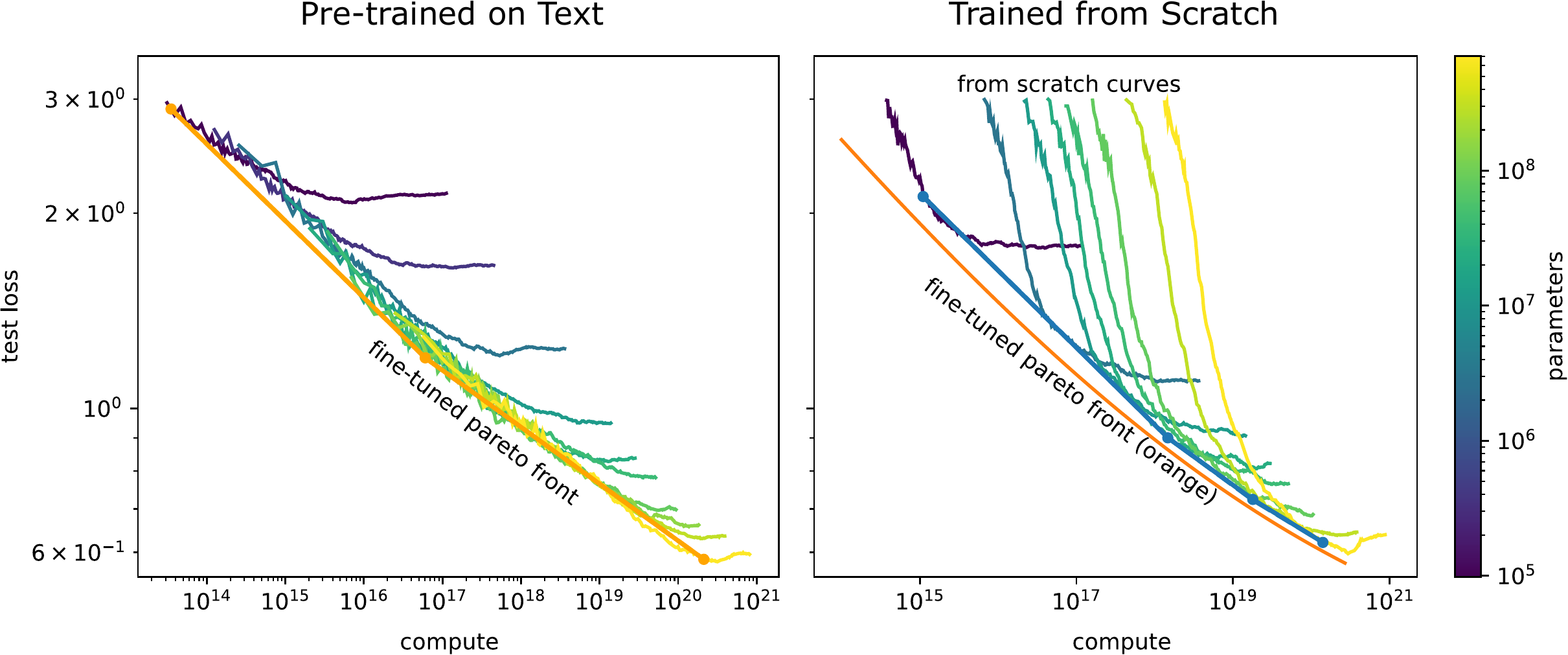}
\caption[Optimal Model Size ]{We show training curves for training on a 10B-character python dataset, parameterized by the amount of compute used for training.
\label{fig:compute_curves_1e10}}
\end{figure}

When we have approximately 30x more data than we had in Figure \ref{fig:compute_curves_3e8}, the compute efficient frontier for fine-tuning is similar to that of from-scratch models. However, it's much easier to be on the
compute efficient frontier when fine-tuning. As shown in Figure \ref{fig:compute_curves_1e10}, the training curves for fine-tuning lie tangent to the frontier for most of training, while the from-scratch curves only lie tangent to the frontier for a relatively narrow window.

However, as shown with the smallest models in Section \ref{section:ossification}, once we start to have
as much or more data than we'd want for training from scratch,
fine-tuning gets substantially worse converged performance and as such
is also less compute efficient.

\begin{figure}[h]
\noindent \centering{} 
\includegraphics[width=1.0\textwidth]{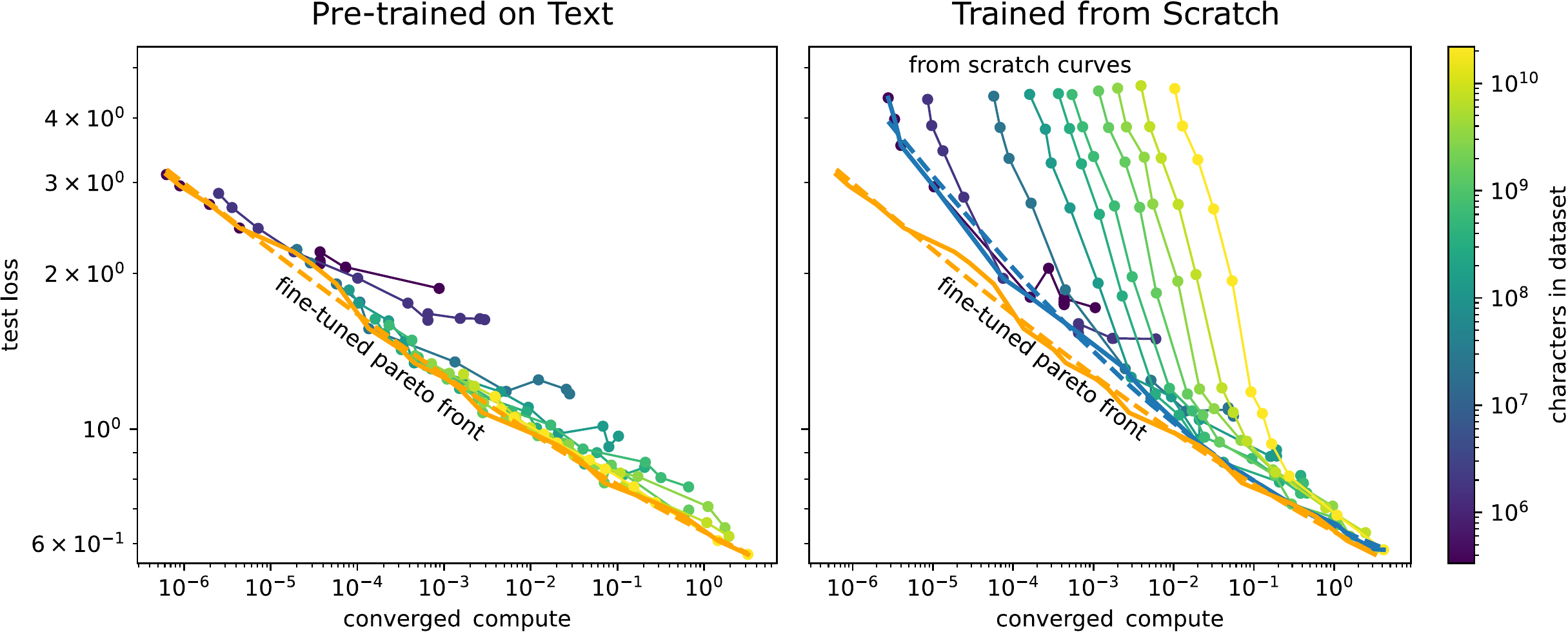}
\caption[Optimal Model Size ]{Different points in a given color here all represent models of a given
size trained to convergence on different sized datasets. An analysis of best epoch, which offers another lens on converged compute costs is given in Appendix \ref{appendix:best_epoch}.
\label{fig:converged_compute}}
\end{figure}

Many models are trained to convergence (compute is used until performance gains stop) rather than the efficient compute
frontier (pareto frontier for performance and compute) \cite{kaplan2020scaling}. In figure \ref{fig:converged_compute} we summarize each curve above with a single point, the converged compute, so we can simultaneously view all the models we trained on datasets of varying size.

When training to convergence: 
\begin{enumerate}
    \item 
Pre-trained models are more compute efficient than training
    from-scratch given a small dataset
    \item 
It's easier to be on the compute
    frontier when fine-tuning as compared to training from scratch, for
    any given dataset size.
\end{enumerate}

\section{Related Work}

Power-laws can arise from a wide variety of sources \cite{thurner2018introduction}. Predictable scaling trends in neural networks were first studied with \cite{1712.00409}. The work closest to our approach is \cite{rosenfeld2019constructive, kaplan2020scaling, henighan2020scaling}. Our focus was on how transfer scales with compute, data, and parameters rather than how performance scales based on those ingredients when training from scratch. 

Transfer and meta-learning have received a lot of attention from the research community in many modalities. We will review some of the work that helped motivate us, but won't do a comprehensive literature review. Here are two recent literature reviews in the domain. \cite{tan2018survey, weng2018metalearning}.

We discussed pre-training language models in the introduction. Pre-training on image datasets such as Instagram and ImageNet has also produced gains in overall performance and data efficiency \cite{mahajan2018exploring, he2019rethinking}. CLIP showed impressive transfer from captioned images by getting zero-shot accuracy comparable to a ResNet-50 on ImageNet on datasets like ImageNet A  \cite{radford2learning, imagenet_cvpr09, hendrycks2021measuring}

Past work in few-shot learning was part of our motivation to study transfer. \cite{Lake1332} showed few-shot learning for generating handwritten characters with probabilistic program induction inline with human capabilities of few-shot learning on such a task. \cite{finn2017modelagnostic} showed that trying to design a model to be fine-tunable can improve few-shot performance. \cite{brown2020language} used existing benchmarks to show that meaningful transfer/few-shot learning can occur in large models on tasks like SuperGLUE \cite{wang2019superglue}. 

Another notable work that helped motivate our investigation into transfer was sim-to-real transfer training for solving a Rubik's cube with a robot hand \cite{openai2019solving}, a setting in which the fine-tuning data is far more expensive than the pre-training data. Another approach for measuring generalization we're interested in is the development of increasingly difficult language benchmarks \cite{hendrycks2021measuring}. 

\section{Limitations}
\begin{enumerate}
\def\labelenumi{\arabic{enumi})}
\item
  Models weren't tuned for fine-tuning or code. We leveraged hyperparameters that were tuned for training from scratch on natural language \cite{ kaplan2020scaling}. We did a handful of learning rate scans for fine-tuning larger models on small dataset sizes and didn't see any improvement with other learning rates. But our scans were not comprehensives.
\item
  Models weren't tuned for small datasets. For small datasets, training ended
  before the warmup was finished, so the learning schedule could
  confound the results.
\item
  We only measured transfer when fine-tuning on python. It's unclear if we'd observe a power law fit for a broad set of distribution pairs.  
\item
  We only measured transfer between distributions in an unsupervised setting. It's not clear to what degree the findings would generalize to a supervised or reinforcement learning setup.
\item
  We didn't find a good closed-form model for from-scratch results as was seen in \cite{kaplan2020scaling}, though we believe more careful tuning could have produced such a model in line with their results. If we had such results we expect we could generate a closed-form equation for overall performance for fine-tuned models rather than rely on relative performance in our definition.
\item
  We only measured performance on transformers.
\item
  Equations 3.1 and 3.2 don't handle zero-shot case unless we use an approximation (i.e. fine-tuning on 1 character) for the zero-shot case.
\item
  We didn't explore the ability to make sure measurements more cheaply, either in context or through KL divergence.

\end{enumerate}

\section{Discussion}

\subsection{Potential unified scaling law for fine-tuning}
\textbf{Using equation 1.5 from \cite{kaplan2020scaling} we find the following result for overall loss on a fine-tuned model in the low-data regime:} 
\begin{equation}
L \approx \left[\left(\frac{N_C}{N}\right)^{\frac{\alpha_N}{\alpha_D}} + \frac{D_C}{k(D_F)^\alpha(N)^\beta}\right]^{\alpha_D} 
\label{eq:unified}
\end{equation}

To generate equation \ref{eq:unified} we have simply substituted effective data from transfer, $D_T$, as given in equation \ref{eq:transfer} for the dataset size $D$ in \cite{kaplan2020scaling}, which was fit to language models trained from scratch. \cite{sharma2020neural} attempts to explain the power-law from \cite{kaplan2020scaling} as arising from neural networks performing regression on a data manifold of intrinsic dimension $d$. Speculatively, we think the scaling laws of transfer fit that picture, where pre-training tiles a portion of the downstream manifold at a lower density than training directly on the downstream task.

\subsection{Speculative estimates for why large models would be few-shot learners}
\label{few-shot}
The effective data multiplier goes down as we increase the size of the fine-tuning dataset, \(D_F\). When we use equation \ref{eq:transfer} to extrapolate all the way down to approximately zero-shot, 1 character, we estimate that pre-trained models are equivalent to training from scratch on 3.7e8 characters of python when pre-trained on text with a model the size of GPT-3 \cite{brown2020language}. If we increase this to a few-shot, say 300 characters, we multiply our effective data \(D_E\) by a factor of 2.8. Given this analysis, it's not surprising that few-shot scores on SuperGLUE \cite{wang2019superglue} were 10-15 points higher than zero-shot scores \cite{brown2020language}. The above analysis is relatively speculative because we extrapolated up in model size by two orders of magnitude and downwards in the amount of data by 5 orders of magnitude. We also equated data in context to data fine-tuned on. 

Similar calculations for text + code pre-training give an estimate of \(D_E\) = 4.1e9 characters, with 300 characters of examples worth a factor 1.7x. We were a bit surprised here, because our best guess before running these experiments would have been that making other programming languages half of the pre-training dataset would have increased few-shot transfer/effective data by more than 10x. One might have the concern that trace amounts of python in the text dataset could impact the transfer coefficients for these experiments. We did additional experiments to mitigate this concern, which are described in Appendix \ref{appendix:confounder}.

\subsection{Potential applications of these scaling laws}
\label{applications}
\begin{enumerate}
\item
    If collecting more data is expensive the power-law form of transfer suggests a potentially useful and cheap experiment when trying to decide whether or not to collect more data to fine-tune a pre-trained model on. One could fine-tune a model with a 1\% and 10\% of the existing dataset. Next one could vary the model size given the full fine-tuning dataset and estimate the lost performance in terms of a reduction in model size. We expect for many applications there will continue to be valuable, expensive to collect data, and that it will be useful to be able to make the decision as to whether to gather more such data in a principled way. One example of an expensive dataset is human preferences of what makes a good summary \cite{stiennon2020learning}.
\item  
    It's easier to generate simple equations for scaling laws where performance is measured in loss rather than accuracy for a downstream task\cite{brown2020language}. Scaling laws posed in terms of effective data could provide an alternative, better behaved trend to compare architectures and algorithms. 

\end{enumerate}

\subsection{Core factors vs details}
\label{core_vs_details}
Previous work showed that the core factors affecting from-scratch performance included data, compute, and parameters \cite{ kaplan2020scaling, henighan2020scaling}. They showed that performance only depended weakly on details like depth, width, and the number of attention heads. We believe fine-tuned performance is likely similar, but where data is split into pre-trained and fine-tuned data. For instance, we expect we'd observe smaller effects for tuning the hyperparameters for the fine-tuning distribution, changing how many epochs we pre-train, and smoothly transitioning between the distributions than for changing the amount of fine-tuning data or the pre-training distribution. We'd be excited to see future work evaluating this claim.

\subsection{How similar are Python and English?}
We believe the dissimilarity of English and Python is representative of transfer between distant distributions that will be of interest in the future. There is English within python code (docstrings, comments, and function names). However, python is a ``formal language'' for communicating instructions to a computer and English is an informal language for communicating between people. We can imagine distributions that are further apart, for instance, English and Math. We can also think of distributions that feel relatively close, like English:French, Wikipedia:arxiv, and so on. Given the distance between distributions of future interest, we'd argue that transfer between text and code tells us something meaningful about transfer between distant distributions.

\subsection{Ossification}
Our small fine-tuned models were unable to reach trained from scratch performance even as they were trained on 10x or 100x $D(N)$, as shown in Figure \ref{fig:ossification}. This is evidence for the intuition some hold that given infinite data one is better off training entirely on distribution and suggests significant pre-training might be impractical in terms of compute and tuning under such circumstances. This suggests that the weights can saturate or ``ossify'', where they become unable to absorb new information well, and that ossification scales predictably. That could be similar to saying that the prior learned in pre-training becomes counterproductive if it's learned with too much relative strength. We may see an analogous phenomena in humans, where there seems to be a large advantage in being trained for a sport from a young age, thus avoiding the opportunity to develop bad habits.

\subsection{Future work we're particularly excited about}
\begin{enumerate}
\item
Measuring transfer coefficients between more unsupervised distributions.
\item
Measuring transfer coefficient where the target distribution is much more general and/or a set of tasks.
\item
Transfer scaling laws where the from-scratch comparisons are better behaved and can generate a unified predictive equation, such as \ref{eq:unified}.
\item
Transfer scaling laws in supervised and reinforcement learning settings.
\item
Transfer scaling laws comparing Transformers to other architectures.
\item
A method to cheaply predict the ideal pre-training ratio for a pair of datasets A and B to maximize performance on a target distribution C, for which we have limited data. It'd be exciting if that ratio were a relatively simple function of the transfer coefficients.
\item  
Given multiple target distributions/downstream tasks, each with limited amounts of data, a set of alpha measurements could potentially enable the construction of an optimal pre-training data split of whatever large datasets would be tractable to put together.
\end{enumerate}

\section{Conclusion}
We've shown that transfer is measurable within language models of a wide range of sizes and that it scales predictably. The units we measure transfer in, data, are intuitive, and given pre-trained models our approach can be used to take such measurements cheaply in the low data regime. We believe our approach is a novel and useful way to understand data as an ML ingredient and the generality of AI systems. We've generated scaling laws for fine-tuning, which has recently become a subject of wide interest. These results help predict the performance, compute, and data needs for scaled-up fine-tuned models.

\section*{Acknowledgments}

We thank Dario Amodei, Jacob Steinhardt, Wojciech Zaremba, Alec Radford, Tom Brown, Alex Ray, Paul Christiano, Amanda Askell, Yura Burda, Ilya Sutskever, Jacob Hilton, Matthias Plappert, and Jakub Pachocki for feedback on this work and numerous helpful conversations.
 
\vfill
\section*{Contributions}
\label{sec:contributions}
\textbf{Danny Hernandez} led the project. He performed the majority of the experiments, analysis, and writing.

\textbf{Jared Kaplan} directly contributed to the analysis and advised the project from its inception.

\textbf{Tom Henighan} maintained the underlying code base and paired on engineering challenges that arose.

\textbf{Sam McCandlish} oversaw the work. In addition to helping to formulate the project and advising on direction, he also generated the dataset, paired on engineering challenges that arose, and directly contributed to the analysis.

\newpage

\bibliographystyle{halpha}
\bibliography{bibliography}

\newcommand{\etalchar}[1]{$^{#1}$}
\begin{thebibliography}{RWC{\etalchar{+}}19}
\expandafter\ifx\csname url\endcsname\relax
  \def\url#1{\texttt{#1}}\fi
\expandafter\ifx\csname doi\endcsname\relax
  \def\doi#1{\burlalt{doi:#1}{http://dx.doi.org/#1}}\fi
\expandafter\ifx\csname urlprefix\endcsname\relax\def\urlprefix{URL }\fi
\expandafter\ifx\csname href\endcsname\relax
  \def\href#1#2{#2}\fi
\expandafter\ifx\csname burlalt\endcsname\relax
  \def\burlalt#1#2{\href{#2}{#1}}\fi

\bibitem[AH18]{AI_And_Compute}
Dario Amodei and Danny Hernandez.
\newblock {AI and Compute}, May 2018.
\newblock \urlprefix\url{https://blog.openai.com/ai-and-compute/}.

\bibitem[BMR{\etalchar{+}}20]{brown2020language}
Tom~B. Brown, Benjamin Mann, Nick Ryder, Melanie Subbiah, Jared Kaplan,
  Prafulla Dhariwal, Arvind Neelakantan, Pranav Shyam, Girish Sastry, Amanda
  Askell, Sandhini Agarwal, Ariel Herbert-Voss, Gretchen Krueger, Tom Henighan,
  Rewon Child, Aditya Ramesh, Daniel~M. Ziegler, Jeffrey Wu, Clemens Winter,
  Christopher Hesse, Mark Chen, Eric Sigler, Mateusz Litwin, Scott Gray,
  Benjamin Chess, Jack Clark, Christopher Berner, Sam McCandlish, Alec Radford,
  Ilya Sutskever, and Dario Amodei.
\newblock Language models are few-shot learners, 2020,
  \burlalt{2005.14165}{http://arxiv.org/abs/2005.14165}.

\bibitem[DCLT18]{1810.04805}
Jacob Devlin, Ming-Wei Chang, Kenton Lee, and Kristina Toutanova.
\newblock Bert: Pre-training of deep bidirectional transformers for language
  understanding, 2018,
  \burlalt{arXiv:1810.04805}{http://arxiv.org/abs/arXiv:1810.04805}.

\bibitem[DDS{\etalchar{+}}09]{imagenet_cvpr09}
J.~Deng, W.~Dong, R.~Socher, L.-J. Li, K.~Li, and L.~Fei-Fei.
\newblock {ImageNet: A Large-Scale Hierarchical Image Database}.
\newblock In {\em CVPR09}, 2009.
\newblock \urlprefix\url{http://www.image-net.org/papers/imagenet_cvpr09.pdf}.

\bibitem[DL15]{dai2015semisupervised}
Andrew~M. Dai and Quoc~V. Le.
\newblock Semi-supervised sequence learning, 2015,
  \burlalt{1511.01432}{http://arxiv.org/abs/1511.01432}.

\bibitem[FAL17]{finn2017modelagnostic}
Chelsea Finn, Pieter Abbeel, and Sergey Levine.
\newblock Model-agnostic meta-learning for fast adaptation of deep networks,
  2017, \burlalt{1703.03400}{http://arxiv.org/abs/1703.03400}.

\bibitem[HB20]{hernandez2020measuring}
Danny Hernandez and Tom~B. Brown.
\newblock Measuring the algorithmic efficiency of neural networks, 2020,
  \burlalt{2005.04305}{http://arxiv.org/abs/2005.04305}.

\bibitem[HBB{\etalchar{+}}21]{hendrycks2021measuring}
Dan Hendrycks, Collin Burns, Steven Basart, Andy Zou, Mantas Mazeika, Dawn
  Song, and Jacob Steinhardt.
\newblock Measuring massive multitask language understanding, 2021,
  \burlalt{2009.03300}{http://arxiv.org/abs/2009.03300}.

\bibitem[HGD19]{he2019rethinking}
Kaiming He, Ross Girshick, and Piotr Doll{\'a}r.
\newblock Rethinking imagenet pre-training.
\newblock In {\em Proceedings of the IEEE/CVF International Conference on
  Computer Vision}, pages 4918--4927, 2019.

\bibitem[HKK{\etalchar{+}}20]{henighan2020scaling}
Tom Henighan, Jared Kaplan, Mor Katz, Mark Chen, Christopher Hesse, Jacob
  Jackson, Heewoo Jun, Tom~B. Brown, Prafulla Dhariwal, Scott Gray, Chris
  Hallacy, Benjamin Mann, Alec Radford, Aditya Ramesh, Nick Ryder, Daniel~M.
  Ziegler, John Schulman, Dario Amodei, and Sam McCandlish.
\newblock Scaling laws for autoregressive generative modeling, 2020,
  \burlalt{2010.14701}{http://arxiv.org/abs/2010.14701}.

\bibitem[HNA{\etalchar{+}}17]{1712.00409}
Joel Hestness, Sharan Narang, Newsha Ardalani, Gregory Diamos, Heewoo Jun,
  Hassan Kianinejad, Md. Mostofa~Ali Patwary, Yang Yang, and Yanqi Zhou.
\newblock Deep learning scaling is predictable, empirically, 2017,
  \burlalt{1712.00409}{http://arxiv.org/abs/1712.00409}.

\bibitem[HR18]{howard2018universal}
Jeremy Howard and Sebastian Ruder.
\newblock Universal language model fine-tuning for text classification, 2018,
  \burlalt{1801.06146}{http://arxiv.org/abs/1801.06146}.

\bibitem[KB14]{kingma2014adam}
Diederik~P. Kingma and Jimmy Ba.
\newblock Adam: A method for stochastic optimization, 2014,
  \burlalt{1412.6980}{http://arxiv.org/abs/1412.6980}.

\bibitem[KMH{\etalchar{+}}20]{kaplan2020scaling}
Jared Kaplan, Sam McCandlish, Tom Henighan, Tom~B. Brown, Benjamin Chess, Rewon
  Child, Scott Gray, Alec Radford, Jeffrey Wu, and Dario Amodei.
\newblock Scaling laws for neural language models, 2020,
  \burlalt{2001.08361}{http://arxiv.org/abs/2001.08361}.

\bibitem[LST15]{Lake1332}
Brenden~M. Lake, Ruslan Salakhutdinov, and Joshua~B. Tenenbaum.
\newblock Human-level concept learning through probabilistic program induction.
\newblock {\em Science}, 350(6266):1332--1338, 2015,
  \burlalt{https://science.sciencemag.org/content/350/6266/1332.full.pdf}{http://arxiv.org/abs/https://science.sciencemag.org/content/350/6266/1332.full.pdf}.
\newblock \doi{10.1126/science.aab3050}.

\bibitem[MGR{\etalchar{+}}18]{mahajan2018exploring}
Dhruv Mahajan, Ross Girshick, Vignesh Ramanathan, Kaiming He, Manohar Paluri,
  Yixuan Li, Ashwin Bharambe, and Laurens van~der Maaten.
\newblock Exploring the limits of weakly supervised pretraining, 2018,
  \burlalt{1805.00932}{http://arxiv.org/abs/1805.00932}.

\bibitem[NKB{\etalchar{+}}19]{nakkiran2019deep}
Preetum Nakkiran, Gal Kaplun, Yamini Bansal, Tristan Yang, Boaz Barak, and Ilya
  Sutskever.
\newblock Deep double descent: Where bigger models and more data hurt, 2019,
  \burlalt{1912.02292}{http://arxiv.org/abs/1912.02292}.

\bibitem[OAA{\etalchar{+}}19]{openai2019solving}
OpenAI, Ilge Akkaya, Marcin Andrychowicz, Maciek Chociej, Mateusz Litwin, Bob
  McGrew, Arthur Petron, Alex Paino, Matthias Plappert, Glenn Powell, Raphael
  Ribas, Jonas Schneider, Nikolas Tezak, Jerry Tworek, Peter Welinder, Lilian
  Weng, Qiming Yuan, Wojciech Zaremba, and Lei Zhang.
\newblock Solving rubik's cube with a robot hand, 2019,
  \burlalt{1910.07113}{http://arxiv.org/abs/1910.07113}.

\bibitem[OBB{\etalchar{+}}19]{openai2019dota}
OpenAI, Christopher Berner, Greg Brockman, Brooke Chan, Vicki Cheung,
  Przemys{\l}aw D{\k e}biak, Christy Dennison, David Farhi, Quirin Fischer,
  Shariq Hashme, Chris Hesse, Rafal J{\'o}zefowicz, Scott Gray, Catherine
  Olsson, Jakub Pachocki, Michael Petrov, Henrique~Pond{\'e} de~Oliveira~Pinto,
  Jonathan Raiman, Tim Salimans, Jeremy Schlatter, Jonas Schneider, Szymon
  Sidor, Ilya Sutskever, Jie Tang, Filip Wolski, and Susan Zhang.
\newblock Dota 2 with large scale deep reinforcement learning.
\newblock 2019, \burlalt{1912.06680}{http://arxiv.org/abs/1912.06680}.
\newblock \urlprefix\url{https://arxiv.org/abs/1912.06680}.

\bibitem[PNI{\etalchar{+}}18]{peters2018deep}
Matthew~E. Peters, Mark Neumann, Mohit Iyyer, Matt Gardner, Christopher Clark,
  Kenton Lee, and Luke Zettlemoyer.
\newblock Deep contextualized word representations, 2018,
  \burlalt{1802.05365}{http://arxiv.org/abs/1802.05365}.

\bibitem[RKH{\etalchar{+}}]{radford2learning}
Alec Radford, Jong~Wook Kim, Chris Hallacy, Aditya Ramesh, Gabriel Goh,
  Sandhini Agarwal, Girish Sastry, Amanda Askell, Pamela Mishkin, Jack Clark,
  et~al.
\newblock Learning transferable visual models from natural language
  supervision.
\newblock {\em Image}, 2:T2.

\bibitem[RNSS18]{radford2018improving}
Alec Radford, Karthik Narasimhan, Tim Salimans, and Ilya Sutskever.
\newblock Improving language understanding by generative pre-training.
\newblock {\em URL https://s3-us-west-2. amazonaws.
  com/openai-assets/research-covers/languageunsupervised/language understanding
  paper. pdf}, 2018.

\bibitem[RRBS19]{rosenfeld2019constructive}
Jonathan~S. Rosenfeld, Amir Rosenfeld, Yonatan Belinkov, and Nir Shavit.
\newblock A constructive prediction of the generalization error across scales,
  2019, \burlalt{1909.12673}{http://arxiv.org/abs/1909.12673}.

\bibitem[RSR{\etalchar{+}}20]{raffel2020exploring}
Colin Raffel, Noam Shazeer, Adam Roberts, Katherine Lee, Sharan Narang, Michael
  Matena, Yanqi Zhou, Wei Li, and Peter~J. Liu.
\newblock Exploring the limits of transfer learning with a unified text-to-text
  transformer, 2020, \burlalt{1910.10683}{http://arxiv.org/abs/1910.10683}.

\bibitem[RWC{\etalchar{+}}19]{radford2019language}
Alec Radford, Jeff Wu, Rewon Child, David Luan, Dario Amodei, and Ilya
  Sutskever.
\newblock Language models are unsupervised multitask learners.
\newblock {\em openai.com}, 2019.

\bibitem[SHM{\etalchar{+}}16]{silver2016mastering}
David Silver, Aja Huang, Chris~J Maddison, Arthur Guez, Laurent Sifre, George
  Van Den~Driessche, Julian Schrittwieser, Ioannis Antonoglou, Veda
  Panneershelvam, Marc Lanctot, et~al.
\newblock Mastering the game of go with deep neural networks and tree search.
\newblock {\em nature}, 529(7587):484--489, 2016.

\bibitem[SK20]{sharma2020neural}
Utkarsh Sharma and Jared Kaplan.
\newblock A neural scaling law from the dimension of the data manifold, 2020,
  \burlalt{2004.10802}{http://arxiv.org/abs/2004.10802}.

\bibitem[SOW{\etalchar{+}}20]{stiennon2020learning}
Nisan Stiennon, Long Ouyang, Jeff Wu, Daniel~M. Ziegler, Ryan Lowe, Chelsea
  Voss, Alec Radford, Dario Amodei, and Paul Christiano.
\newblock Learning to summarize from human feedback, 2020,
  \burlalt{2009.01325}{http://arxiv.org/abs/2009.01325}.

\bibitem[SSS{\etalchar{+}}17]{Silver_2017}
David Silver, Julian Schrittwieser, Karen Simonyan, Ioannis Antonoglou, Aja
  Huang, Arthur Guez, Thomas Hubert, Lucas Baker, Matthew Lai, Adrian Bolton,
  and et~al.
\newblock Mastering the game of go without human knowledge.
\newblock {\em Nature}, 550(7676):354--359, Oct 2017.
\newblock \doi{10.1038/nature24270}.

\bibitem[THK18]{thurner2018introduction}
Stefan Thurner, Rudolf Hanel, and Peter Klimek.
\newblock {\em Introduction to the theory of complex systems}.
\newblock Oxford University Press, 2018.

\bibitem[TSK{\etalchar{+}}18]{tan2018survey}
Chuanqi Tan, Fuchun Sun, Tao Kong, Wenchang Zhang, Chao Yang, and Chunfang Liu.
\newblock A survey on deep transfer learning.
\newblock In {\em International conference on artificial neural networks},
  pages 270--279. Springer, 2018.

\bibitem[VSP{\etalchar{+}}17]{OriginalTransformer}
Ashish Vaswani, Noam Shazeer, Niki Parmar, Jakob Uszkoreit, Llion Jones,
  Aidan~N Gomez, \L~ukasz Kaiser, and Illia Polosukhin.
\newblock Attention is all you need.
\newblock In I.~Guyon, U.~V. Luxburg, S.~Bengio, H.~Wallach, R.~Fergus,
  S.~Vishwanathan, and R.~Garnett, editors, {\em Advances in Neural Information
  Processing Systems 30}, pages 5998--6008. Curran Associates, Inc., 2017.
\newblock
  \urlprefix\url{http://papers.nips.cc/paper/7181-attention-is-all-you-need.pdf}.

\bibitem[Wen18]{weng2018metalearning}
Lilian Weng.
\newblock Meta-learning: Learning to learn fast.
\newblock {\em lilianweng.github.io/lil-log}, 2018.
\newblock
  \urlprefix\url{http://lilianweng.github.io/lil-log/2018/11/29/meta-learning.html}.

\bibitem[WPN{\etalchar{+}}19]{wang2019superglue}
Alex Wang, Yada Pruksachatkun, Nikita Nangia, Amanpreet Singh, Julian Michael,
  Felix Hill, Omer Levy, and Samuel~R. Bowman.
\newblock Superglue: A stickier benchmark for general-purpose language
  understanding systems, 2019,
  \burlalt{1905.00537}{http://arxiv.org/abs/1905.00537}.

\end{thebibliography}

\newpage
\appendix

\section{Data Regime}
\label{appendix:data_regime}

To define data regimes we estimate \(D(N)\), the amount of data it takes to get 99\% of the performance that infinite python data would yield for a given model size. \(D(N)\) approximately defines the "infinite data regime" as a function of model size. We consider \(D_F \leq 10\%\) of \(D(N)\) to be the low data regime.

\begin{figure}[h]
\noindent \centering{} 
\includegraphics[width=.70\textwidth]{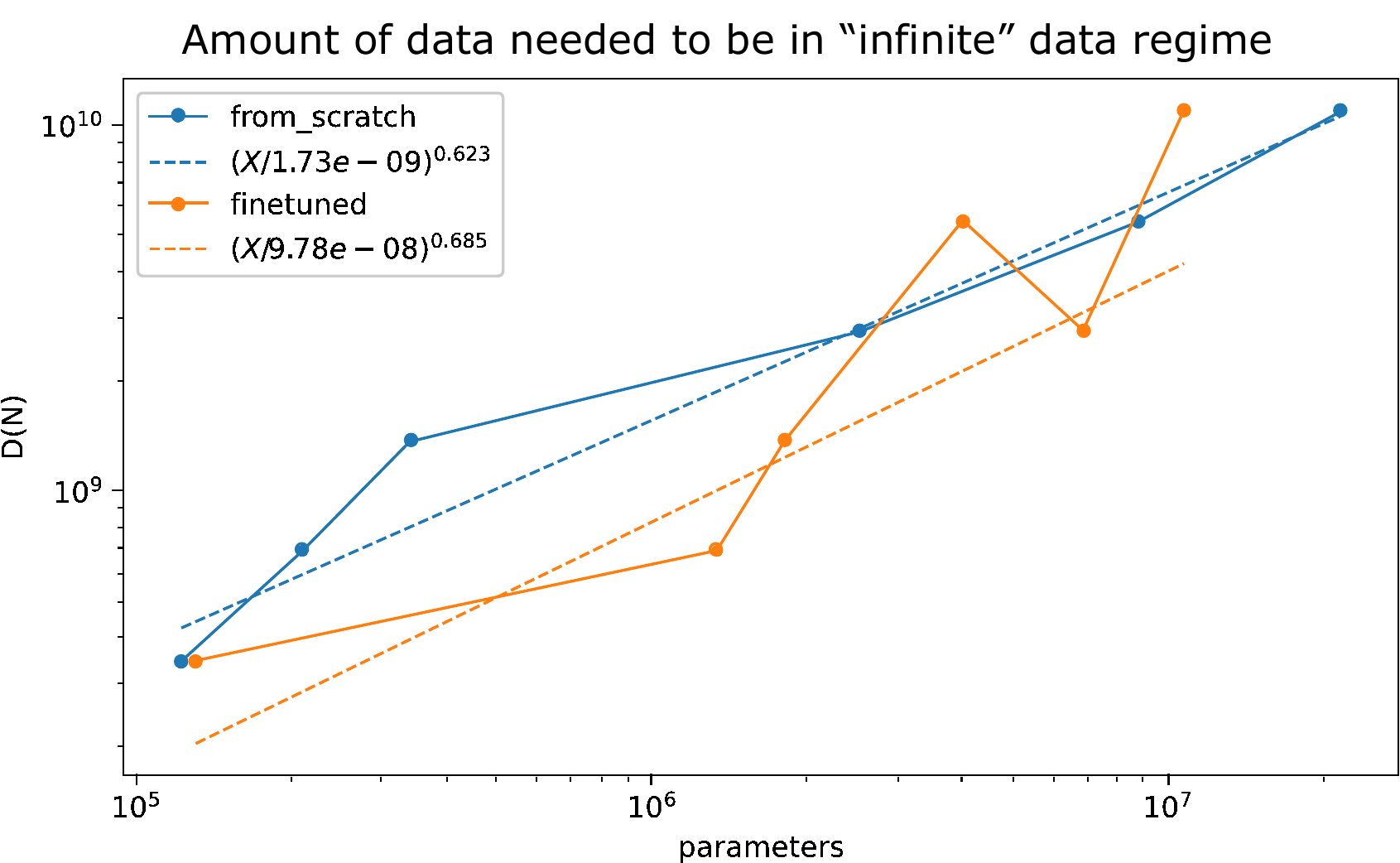}
\caption[Optimal Model Size ]{\textbf{Estimating python data needs}
\label{fig:D_N}}
\end{figure}

D(N) was calculated by determining where curves in figure \ref{fig:scaling_laws} with a given fraction of the dataset intersected curves with the full dataset. Intersecting was defined by 99\% performance for from-scratch and 95\% for fine-tuned. The difference was a judgment call, made based on final fit and the fine-tuning intersections looking relatively noisier.

\section{Supplementary equations}

The relation between total effective data, $D_E$, effective data from transfer, $D_T$, and fine-tuning dataset $D_F$ is shown visually in Figure \ref{fig:visual_explanation}. It's also shown below for clarity.

\begin{equation}
\textrm{total effective data} = D_E = D_T + D_F 
\label{eq:effective_data}
\end{equation}

In Figure \ref{fig:effective_data} the vertical axis is the fraction of effective data from transfer. We give the explicit equation for it below in terms of equation \ref{eq:transfer} and \ref{eq:effective_data}:

\begin{equation}
    \textrm{fraction effective data from transfer} = \frac{D_T}{D_F+D_T} =\frac{k(D_F)^{\alpha-1}(N)^\beta}{1+k(D_F)^{\alpha-1}(N)^\beta} 
\label{eq:fraction}
\end{equation}

\section{Generating fit for Equation 1.2}
\label{appendix:fit}

\begin{figure}[h]
\noindent \centering{} 
\includegraphics[width=1.0\textwidth]{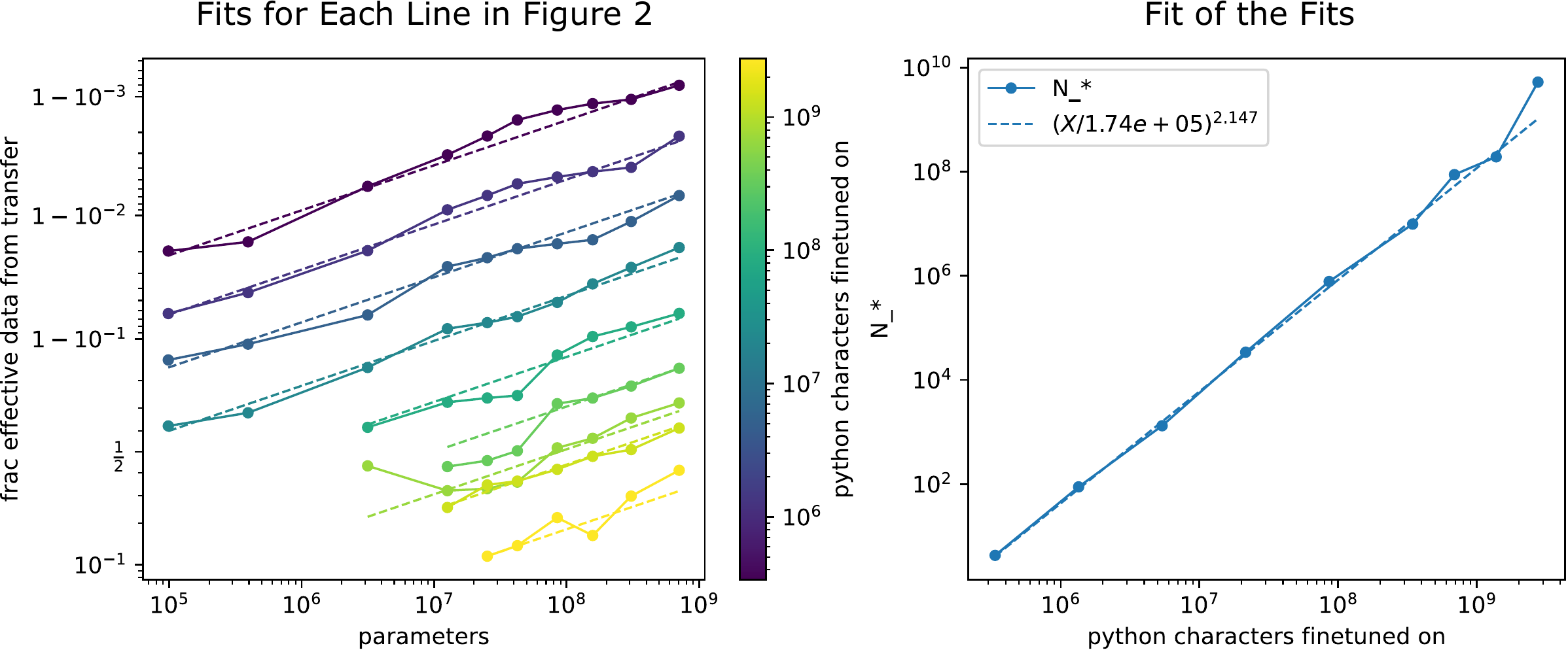}
\caption[Optimal Model Size ]{When we noticed how straight the lines were with individual fits on the left, 
\label{fig:n_star}}
\end{figure}

The good fit we found to our data grouped by \(D_F\) is shown on the left in Figure \ref{fig:n_star}.

We sought out a global fit for our experiments with the goal of generating more insights and increasing predictive capability. What ended up working was fitting the fits. We noticed that the logit fits on the right of \ref{fig:n_star} all had approximately the same exponent. So we tried fitting the following equation to the fits.

\begin{equation}
\textrm{fraction effective data from transfer} = \frac{D_T}{D_F+D_T} = \frac{1}{1+\left(\frac{N*}{N}\right)^{.38}}
\end{equation}

The fit to \(N_*\) is then used to generate the global fit shown on the left of figure \ref{fig:effective_data} equation \ref{eq:transfer}. This fit gives equal weight to each dataset size, \(D_F\), that we fine-tuned on. A similar approach was used for generating the fit on the right side of the equation.

Before coming across this fit of fits on the logits of the fraction of effective data from transfer we attempted to find other regularities in the data. We plotted effective data on it's own.

\begin{figure}[h]
\noindent \centering{} 
\includegraphics[width=.50\textwidth]{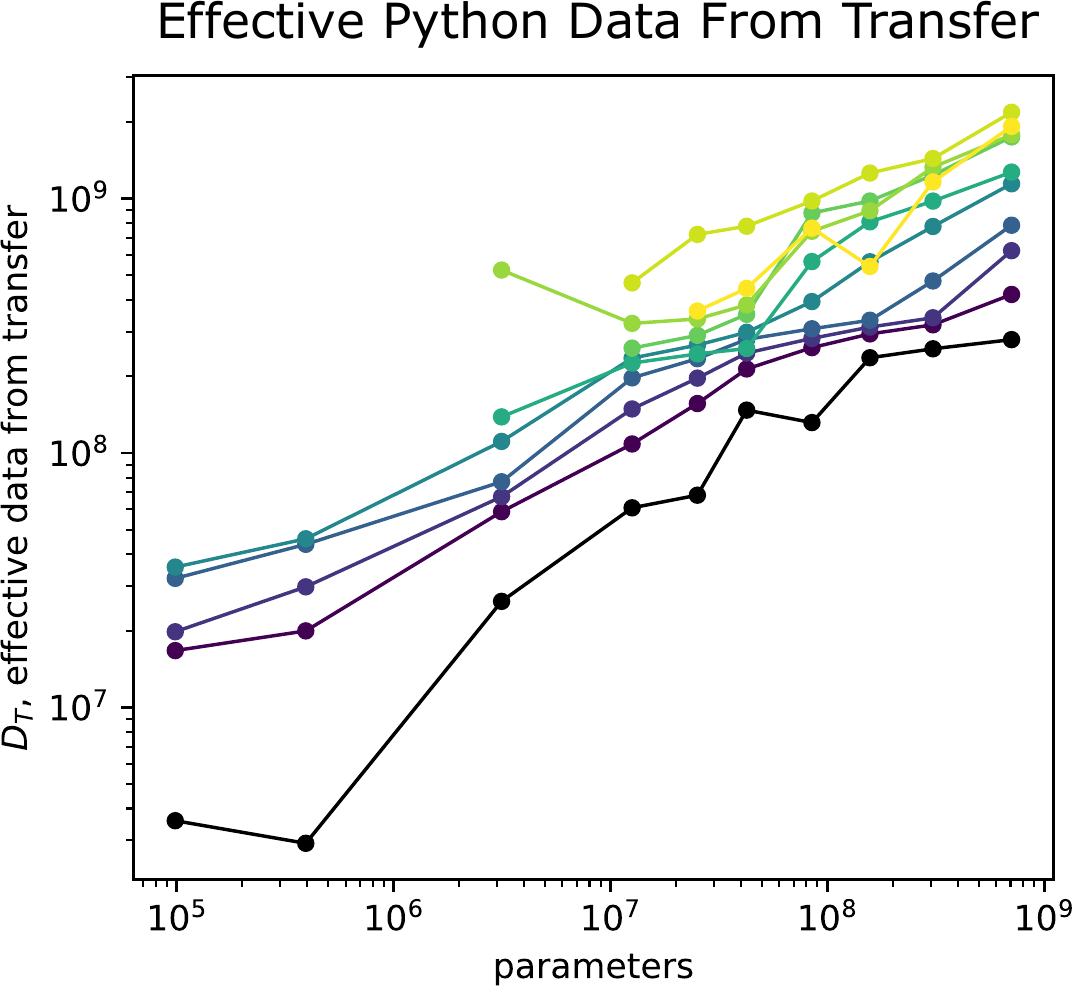}
\caption[Optimal Model Size ]{Initial plots of effective data were promising, in that it seemed to yield somewhat noisy   lines in log space.
\label{fig:effective_data_appendix}}
\end{figure}

We also looked for patterns in $D_T/D_F$. When we normalized $D_T/D_F$ with $D(N)$ as shown in Figure \ref{fig:ossification} we found the results relatively promising. However, once we saw the straights lines for $D_T/D_E$ when plotted with logit axes, we focused entirely on that line of analysis.

\begin{figure}[h]
\noindent \centering{} 
\includegraphics[width=.60\textwidth]{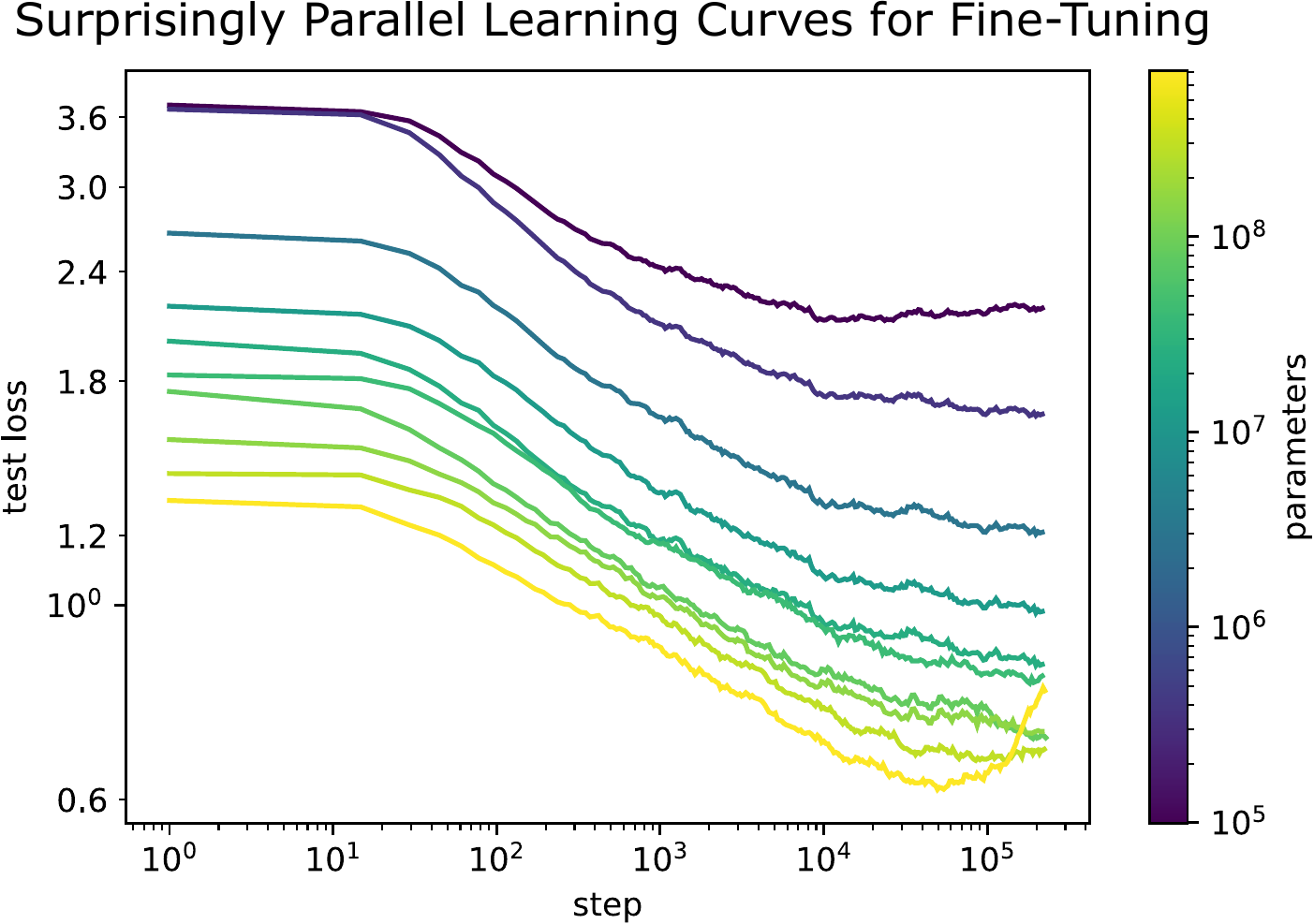}
\caption[Optimal Model Size ]{The above models were fine-tuned on 5.5e9 python characters.
\label{fig:learning_curves}}
\end{figure}

We see that the largest model starts to obviously overfit near the end of training. Aside from that, the learning curves are relatively parallel. We attempted to use this to make predictions of quantitative performance but found the predictions to be too inaccurate to be worthwhile. However, we still find the qualitative prediction that fine-tuning learning curves should be very parallel as you scale up to be somewhat useful. 

\newpage
\section{Figure 3, including medium data regime}
\label{appendix:omitted}

\begin{figure}[h]
\noindent \centering{} 
\includegraphics[width=\textwidth]{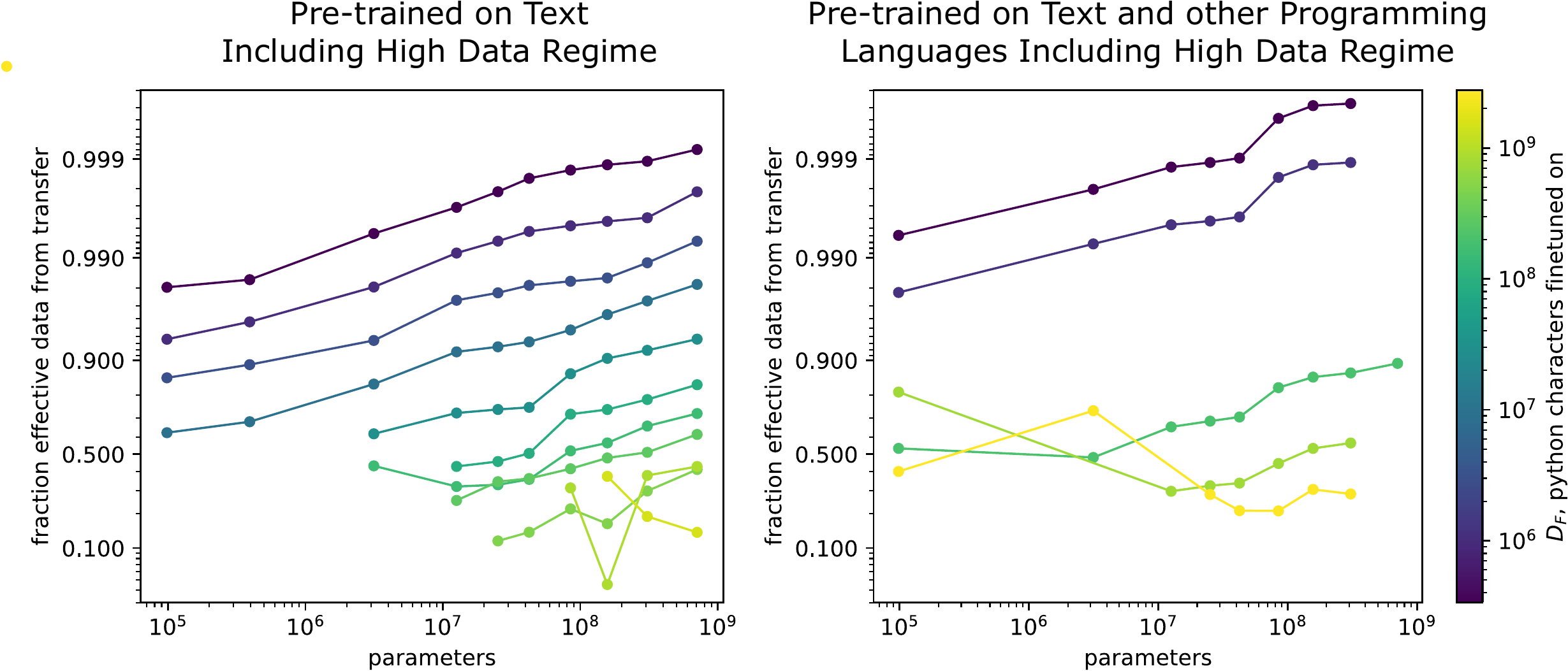}
\caption[Optimal Model Size ]{We no longer get a good fit for these models once we leave the low-data regime. We don't see as large of a breakdown in fit for models pre-trained on text only. We still only show points for which $D_T > 0$.
\label{fig:omitted}}
\end{figure}

We observe in figure \ref{fig:omitted} below that the fit given by equation 3.2 breaks down for transfer in these distributions in the medium data regime \(D(N) > .10\).

We suspect that the behavior is less regular (no longer can be fit to a straight line) in the high data regime for two reasons:
\begin{enumerate}
    \item 
    The estimation of effective data through interpolation is poorly conditioned in the high data regime. The curves in the high data regime in Figure \ref{fig:scaling_laws} lie nearly on top of each other so small variations in model performance could have a large effect.
    \item
    As we discuss in Section \ref{core_vs_details}, we believe tuning matters more in the high data regime, and we did relatively little tuning.
    \end{enumerate}
    
\section{Attempt to generate global fit to our from-scratch python runs}
We attempted to fit global power-laws to the from-scratch python runs, as was done for language models in \cite{kaplan2020scaling} and in more modalities in \cite{henighan2020scaling}. Specifically, we tried to fit Equation 1.5 from \cite{kaplan2020scaling} to runs shown on the left side of Figure \ref{fig:scaling_laws}.
 
\begin{equation}
L \approx \left[\left(\frac{N_C}{N}\right)^{\frac{\alpha_N}{\alpha_D}} + \frac{D_C}{D}\right]^{\alpha_D} 
\label{eq:original_scaling_law}
\end{equation}

\begin{figure}[h]
\noindent \centering{} 
\includegraphics[width=1.0\textwidth]{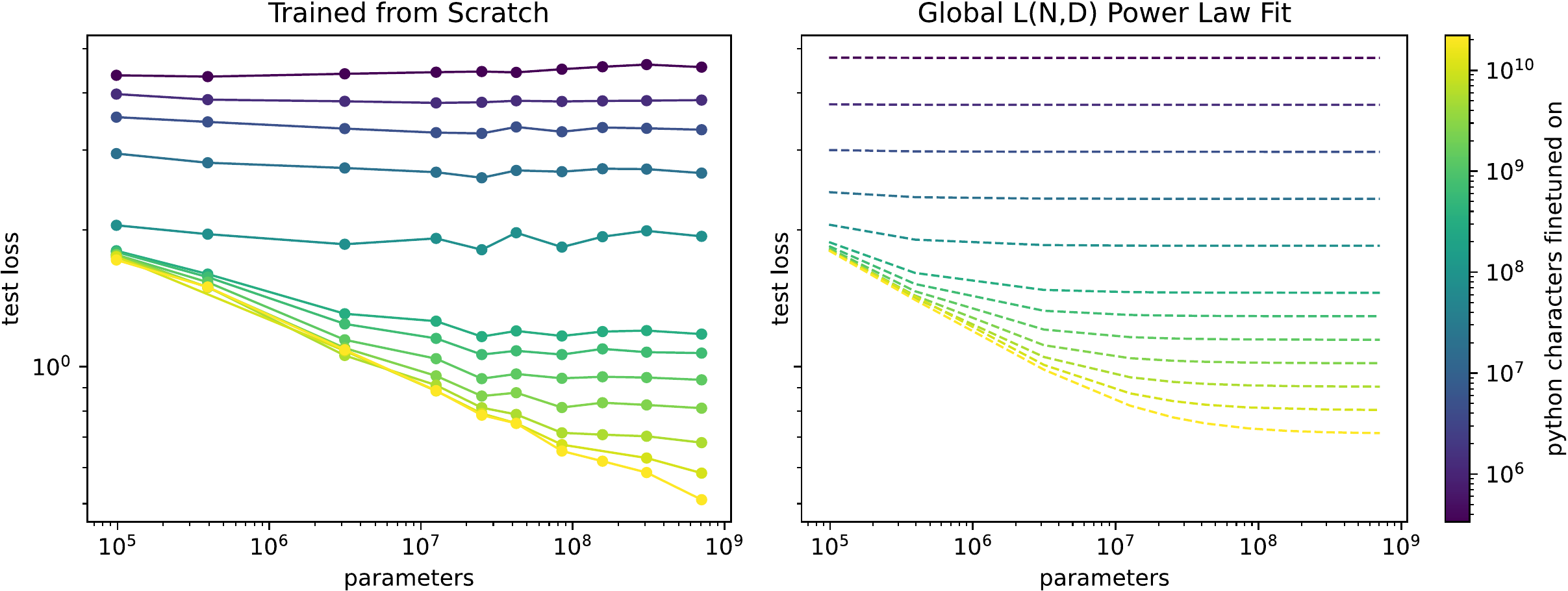}
\caption[Optimal Model Size ]{The above fit for models trained on python from-scratch wasn't a good enough to use as foundation to layer on an additional fit for fine-tuning.
\label{fig:scratch_fits}}
\end{figure}

It seems plausible to us that more careful tuning of these experiments would have generated a good fit here, but the fit was too poor to serve as a foundation for a unified scaling law. It may also be the case that a modified version of \ref{eq:original_scaling_law} of the type proposed in \cite{rosenfeld2019constructive} would be more appropriate. By unified scaling law we mean a function $L(N,D_F)$ for fine-tuning on python with the same coefficients as an $L(N,D)$ fit as shown in equation \ref{eq:original_scaling_law} for python models trained from scratch.

\section{Addressing potential concern about small amounts of python in text dataset}
\label{appendix:confounder}

We ran additional experiments to show that the majority of the zero-shot transfer discussed in Section \ref{few-shot} was the result from "transfer" from text to python. The zero-shot performance for our larger models was similar to what we'd expect if we trained from scratch on an amount of python equal to about .3\% of our text dataset in size. Rather than attempt to measure the amount of python in the test dataset, we added a known amount of python, 0.3\%, into pre-training. If we were simply measuring the amount of python in our natural language dataset, we would have expected the effective amount of python data to go up by a factor of 2x. Instead, mixing in the 0.3\% increased the effective python data by a factor of 4.5x. As a result of this experiment, we conclude that the majority of what we measured in the previous graphs is in fact "transfer" from a text distribution to a python distribution rather than the uncontrolled amount of python in the pre-training distribution. 

\newpage
\section{Analysis of best epoch}
\label{appendix:best_epoch}
An important practical question for fine-tuning is how long one will need to train for before network performance saturates. The short answer is, if the dataset is quite large, a similar length of time as it'd take from-scratch: 1-10 epochs for code, where epochs generally go down with model size and increased data. 

\begin{figure}[h]
\noindent \centering{} 
\includegraphics[width=1.0\textwidth]{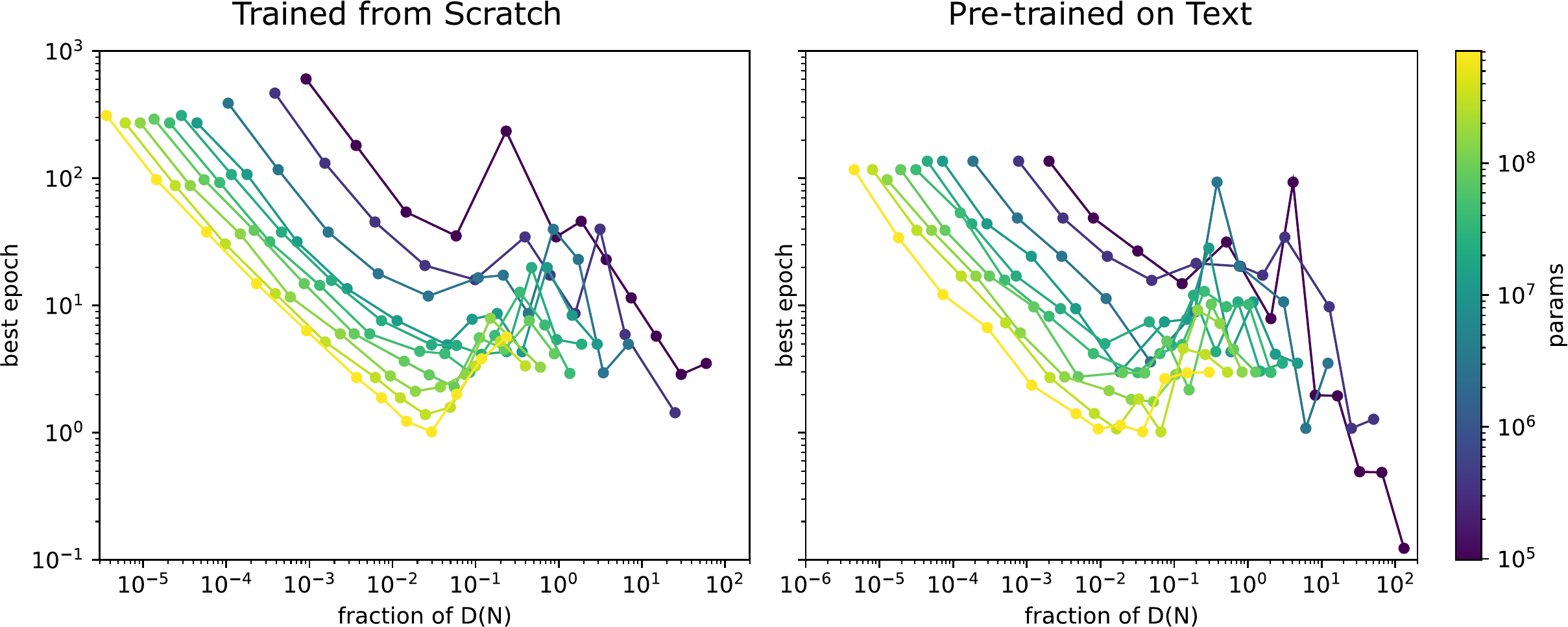}
\caption[Optimal Model Size ]{Finetuning on significant amounts of data requires about the same number of epochs as training from scratch, whereas finetuning on a small data requires 2-5x fewer epochs. Parametrizing in terms of $D(N)$ makes the patterns appear cleaner.
\label{fig:best_epoch}}
\end{figure}

If the fine-tuning dataset is small, it's easy to just train it and do early stopping when it begins to obviously overfit. For our smallest dataset (100,000 tokens) it took 2-5x fewer epochs than it'd take from-scratch. The Z-shaped curves are a bit surprising, and are somewhat similar in shape to what was observed in \cite{nakkiran2019deep}. Another limitation here is that for small amounts of data much of the learning happens during warmup, and so the learning rate schedule confounds the optimum number of epochs.

\end{document}